\documentclass[journal]{IEEEtran}
\usepackage{graphicx}
\usepackage{subfigure}
\usepackage{array}
\usepackage{amsmath}
\usepackage{amssymb}
\usepackage{multirow}
\usepackage{cases}
\usepackage{url}
\usepackage{cite}
\usepackage{fancyhdr}
\usepackage{mathrsfs}
\usepackage[justification=centering]{caption}
\usepackage[colorlinks=true,citecolor=black,linkcolor=black,anchorcolor=black,urlcolor=black,bookmarks=false]{hyperref}
\usepackage{epstopdf}
\usepackage{amsmath}

\ifCLASSINFOpdf

\else

\fi

\begin{document}

\title{Perceptual Depth Quality Assessment of Stereoscopic Omnidirectional Images}

\author{Wei Zhou,~\IEEEmembership{Senior Member,~IEEE}, and Zhou Wang,~\IEEEmembership{Fellow,~IEEE}
\thanks{W. Zhou is with the School of Computer Science and Informatics, Cardiff University, Cardiff CF24 4AG, United Kingdom (e-mail: zhouw26@cardiff.ac.uk).}
\thanks{Z. Wang is with the Department of Electrical and Computer Engineering, University of Waterloo, Waterloo, ON N2L 3G1, Canada (e-mail: zhou.wang@uwaterloo.ca).}
}

\markboth{IEEE Transactions on Circuits and Systems for Video Technology}
{Shell \MakeLowercase{\textit{et al.}}: Bare Demo of IEEEtran.cls for IEEE Journals}

\maketitle

\begin{abstract}
Depth perception plays an essential role in the viewer experience for immersive virtual reality (VR) visual environments. However, previous research investigations in the depth quality of 3D/stereoscopic images are rather limited, and in particular, are largely lacking for 3D viewing of 360-degree omnidirectional content. In this work, we make one of the first attempts to develop an objective quality assessment model named depth quality index (DQI) for efficient no-reference (NR) depth quality assessment of stereoscopic omnidirectional images. Motivated by the perceptual characteristics of the human visual system (HVS), the proposed DQI is built upon multi-color-channel, adaptive viewport selection, and interocular discrepancy features. Experimental results demonstrate that the proposed method outperforms state-of-the-art image quality assessment (IQA) and depth quality assessment (DQA) approaches in predicting the perceptual depth quality when tested using both single-viewport and omnidirectional stereoscopic image databases. Furthermore, we demonstrate that combining the proposed depth quality model with existing IQA methods significantly boosts the performance in predicting the overall quality of 3D omnidirectional images.
\end{abstract}


\IEEEpeerreviewmaketitle

\section{Introduction}

\IEEEPARstart{T}HERE has been a rapid development of virtual reality (VR) technology and a growing popularity of VR devices in recent years. A great amount of VR content has been acquired, stored, transmitted, and displayed in the form of 360-degree omnidirectional images (OIs)~\cite{xi2017virtual}. These OIs may be rendered to cover the whole $180\times {{360}^{{}^\circ }}$ range on a spherical scene surrounding the viewer, and thus provide the viewer with a richer immersive quality-of-experience (QoE) in a 3D environment as compared to conventional 2D images that only occupy a restricted plane \cite{zhu2021viewing,xu2020state}. Beyond the traditional OIs that consist of a single view, it is also possible to render the spherical scene with stereoscopic pairs of OIs with both left and right views, creating even richer and more realistic immersive 3D plus 360-degree viewer experiences \cite{zhou2021projection}. Nevertheless, a variety of quality issues may arise in the creation, transmission and display processes of OIs \cite{azevedo2019visual}. In order for the device manufacturers and service providers of omnidirectional content to optimize the perceptual QoE of end consumers, there is an urgent need of accurate and easy-to-use omnidirectional image quality assessment (OIQA) and 3D OIQA methods \cite{wang2002image,diemer2015impact,xu2018assessing,tian2022vsoiqe}.

In general, OIQA approaches are divided into two main categories $-$ subjective and objective quality assessment. Subjective quality assessment is considered the most accurate method \cite{xu2021perceptual}. Several subjective quality databases for OIs have been built in recent years \cite{sun2018large,duan2018perceptual,li2019cross}, where each OI is associated with a perceptual visual quality (i.e., image quality). For 3D OIs, Chen et al. \cite{chen2019study} constructed the LIVE 3D VR database, where six distortion types and five distortion levels were introduced to generate 450 distorted 3D OIs. The NBU-SOID subjective quality database \cite{qi2020viewport} considered three kinds of classic compression artifacts. Head-mounted displays (HMDs) were adopted in a single stimulus quality evaluation setup and the 3D OIs underwent both symmetric and asymmetric distortions. Besides image quality, depth perception is another essential quality factor \cite{series2015subjective} that affects the overall user experience of 3D OIs \cite{zhou20163d}. This motivated Xu et al. \cite{xu2018subjective} to create the SOLID subjective quality database, in which each 3D OI was rated by three quality dimensions $-$ image quality, depth perception, and overall QoE.

In practice, subjective quality evaluation is often laborious, expensive, and inconvenient \cite{huynh2010study}. Objective quality assessment provides an attractive alternative that not only predicts perceptual quality automatically, but may also be integrated into working VR processing systems for optimal performance. In the literature, there have emerged a number of objective OIQA models \cite{xu2020blind,zhou2021no,fu2022adaptive}. Most of these methods have targeted at image quality, but none of them assesses depth perception of 3D OIs, which is an essential quality dimension in 3D visual QoE. In this work, we focus on designing an effective depth quality measure of 3D OIs, and explore the impact of depth quality assessment to the overall QoE of 3D OIs. To the best of our knowledge, this is the first attempt to address this challenging problem.

\begin{figure*}[t]
	\centering
	\includegraphics[width=18cm]{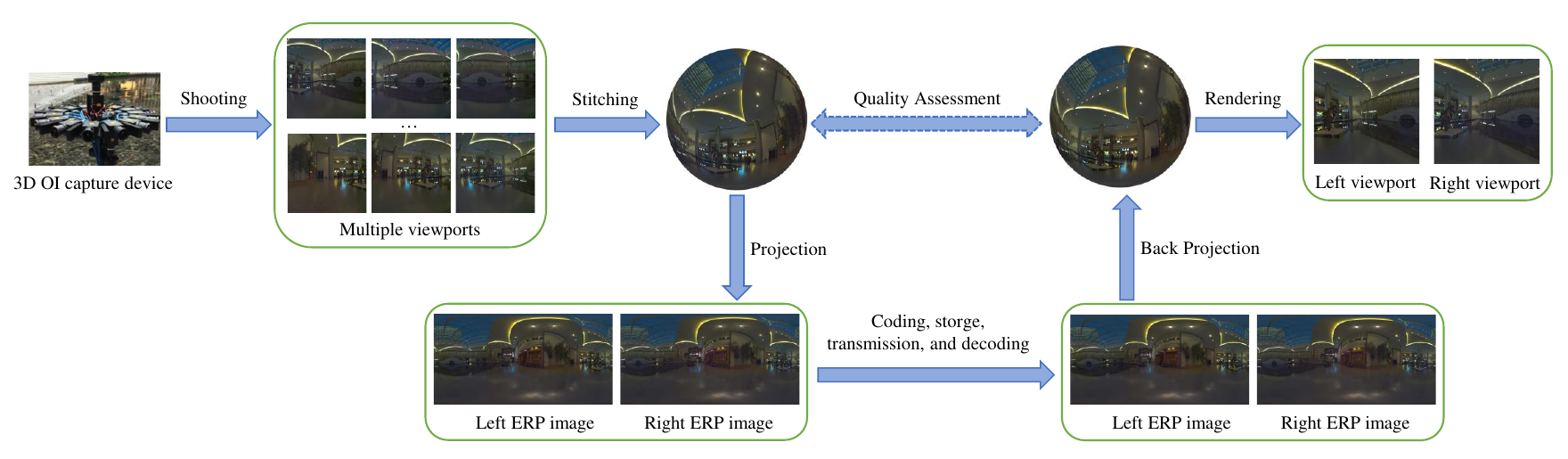}
	\caption{An end-to-end 3D omnidirectional content processing pipeline.}
	\label{fig1}
\end{figure*}

From the perspective of computational models, unlike IQA that is often based on the characteristics of viewed images such as structure, texture and content \cite{wang2006modern}, depth quality assessment (DQA) is harder to deal with, mainly because of the complex 3D vision mechanisms for the binocular human visual system (HVS), especially in the more immersive case of 3D OIs. More importantly, an efficient DQA model should be beneficial to overall QoE prediction. According to the study of psychophysics and neuroscience, binocular depth depends on the ability of the HVS to precisely match corresponding feature representations in the left and right eyes \cite{ohzawa1990stereoscopic,neri1999probing,parker2007binocular,henriksen2016disparity}. Therefore, we employ the interocular discrepancy of left and right views. According to the proposed multi-color-channel and adaptive viewport selection scheme, a blind/no-reference depth quality index (DQI) is designed for assessing the depth quality of both single-viewport and omnidirectional stereoscopic images. In addition to the depth quality evaluation, we further extend the depth quality measure to the modeling of the overall QoE of 3D OIs.

The main contributions of this work are summarized as follows:

\begin{itemize}
\item The interocular discrepancy is proved to be a good depth quality indicator, which can also save computational complexity. The adaptive viewport and region selection approaches are proposed based on the discriminative depth information of interocular discrepancy, for 3D OIs and common 3D images, respectively.
\item Motivated by the perceptual color peculiarity of the HVS, we resort to using color decomposition for interocular discrepancy maps, which is more consistent with the human perception. Based on the depth perception reflecting global properties of input signals and frequency independent mechanism, the interocular discrepancy statistics of decomposed frequency subbands are extracted for the final depth quality regression.
\item We extend the proposed DQI to overall quality assessment of 3D OIs, leading to depth-guided overall QoE measure. We find that by integrating our DQI into classical IQA features, the performance of overall quality prediction for 3D OIs can be significantly boosted, which further demonstrates the effectiveness of our proposed depth quality measure.
\end{itemize}

The rest of this paper is organized as: In Section II, we provide the review of related works, including objective IQA as well as DQA methods for both traditional and omnidirectional images, and then give the motivations of our work. The proposed depth quality measure is described in Section III. Extensive experimental results are presented in Section IV, and we conclude the paper in Section V.

\section{Related Works and Motivations}
In this section, the related objective quality assessment methods are first reviewed, including 2D IQA, OIQA, 3D IQA, 3D OIQA, and DQA models. Then, we present the motivations of our proposed DQI and the depth-guided overall QoE measure.

\subsection{Related Objective Quality Assessment Models}
Since humans are the ultimate receivers of most visual signals, the goal of objective quality assessment is to predict the human-perceived quality. Therefore, the average quality ratings from subjective tests are usually adopted for creating the ground-truth labels for image quality, typically in the form of the mean opinion score (MOS) or difference mean opinion score (DMOS). Based on the availability of original reference images, the objective quality assessment methods generally have three types that consist of full-reference (FR), reduced-reference (RR), and no-reference (NR) quality assessment which is also known as blind quality assessment. The NR models are the most applicable and challenging approaches in real-world, but due to the fully accessible pristine information, FR models could deliver better performance compared with the others \cite{bosse2017deep}. The earliest PSNR calculates the signal fidelity by pixel-to-pixel error. However, this way is inconsistent with human perception. Thus, according to the characteristics of the HVS, the structure similarity (SSIM) \cite{wang2004image} and several variants have been proposed, such as multiscale SSIM (MS-SSIM) \cite{wang2003multiscale} and feature similarity (FSIM) \cite{zhang2011fsim}, etc.

Intuitively, these 2D IQA metrics can be directly applied to the perceptual quality assessment of OIs by performing the computation process on equi-rectangular projection (ERP) format \cite{ERP}. But the ERP images often contain inevitable geometric deformation. To bridge the gap between conventional 2D IQA approaches and omnidirectional characterizations, the spherical PSNR (S-PSNR) \cite{yu2015framework} was proposed to compute PSNR by sampling points on the sphere instead of ERP images. Besides, the weighted-to-spherically uniform PSNR (WS-PSNR) \cite{sun2017weighted} was developed according to the assigned weights of stretching areas. Zakharchenko et al. \cite{zakharchenko2016quality} proposed the craster parabolic projection PSNR (CPP-PSNR), aiming to calculate the PSNR on the CPP plane.

Compared to 2D image data, a 3D image is composed of two 2D images, i.e., left and right views, which brings more challenges to the objective quality assessment of 3D images. For example, if left and right views have different types and degrees of artifacts, asymmetric distortion happens and it is more difficult to evaluate the asymmetrically distorted 3D images \cite{lin2014quality}. It is obvious that directly averaging the predicted quality values from left and right views cannot reveal the binocular mechanisms of the HVS \cite{campisi2007stereoscopic,benoit2009quality}. Therefore, several 3D IQA methods that take specific 3D characteristics into account have been proposed, e.g., the cyclopean \cite{chen2013full} model and weighted SSIM (W-SSIM) as well as weighted FSIM (W-FSIM) \cite{wang2015quality}. For 3D OIQA, a multi-viewport based quality assessment model was proposed \cite{xu2019quality}. Furthermore, the stereoscopic omnidirectional image quality evaluator (SOIQE) \cite{chen2020stereoscopic} was developed on the basis of predictive coding theory.

The above-mentioned 3D quality assessment methods are all designed for estimating image quality rather than depth quality which is of great significance in 3D images. Hence, to tackle this problem, the depth perception difficulty index (DPDI) was presented with a prediction model \cite{wang2016perceptual}. Moreover, in \cite{chen2017blind}, a depth perception quality metric (DPQM) was proposed. Nevertheless, they are depth perception evaluation criteria for 3D images rather than 3D OIs. In other words, there is a lack of DQA method specifically designed for 3D OIs.

\subsection{Motivations}

\begin{figure}[t]
	\centering
	\includegraphics[width=9cm]{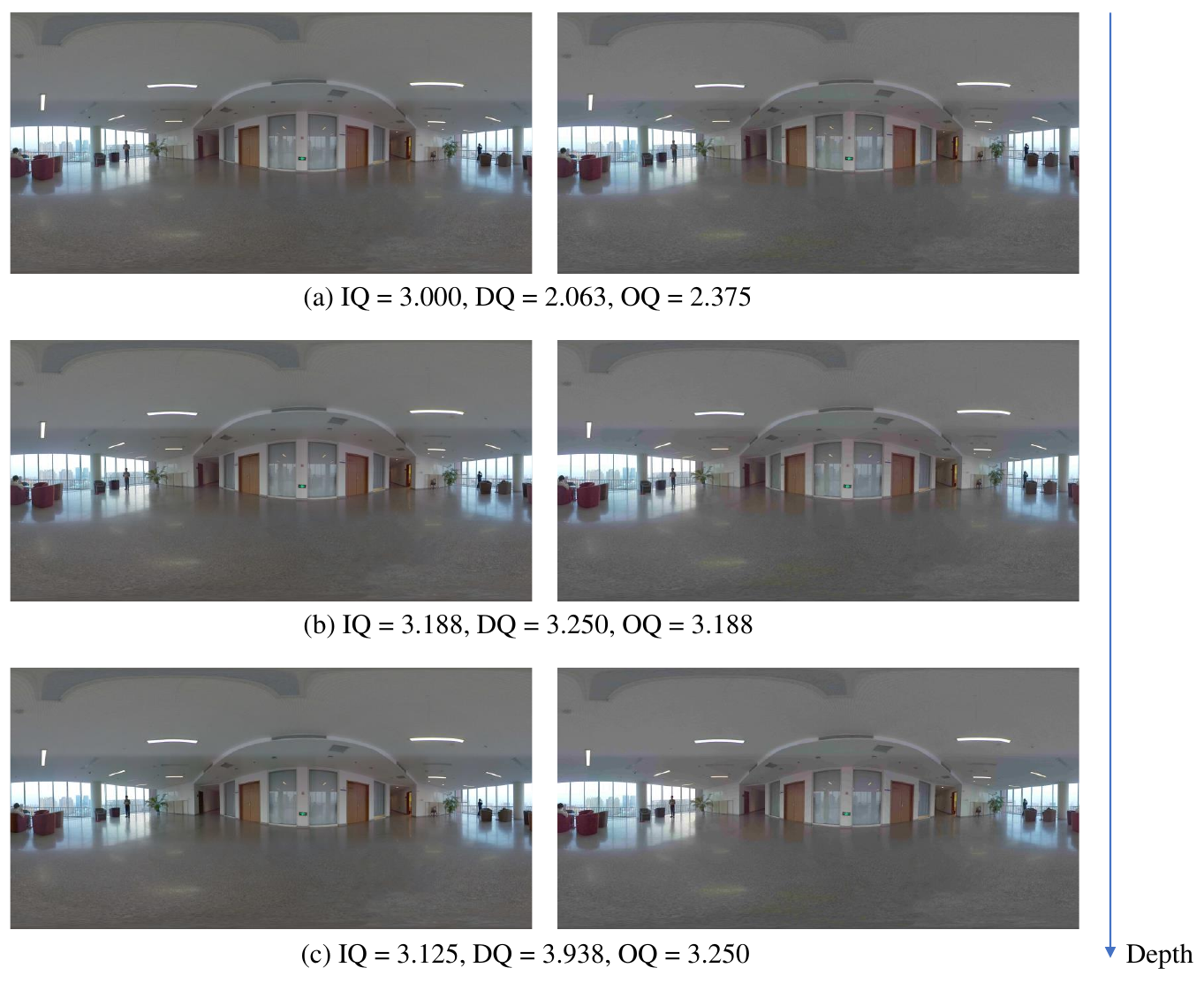}
	\caption{Examples of JPEG compressed 3D OIs with left and right ERP images. Both left and right views have the same compression level separately (level 1 versus level 4). Subjective ground-truth labels are provided. IQ: image quality, DQ: depth quality, OQ: overall QoE.}
	\label{fig2}
\end{figure}

Typically, an end-to-end 3D omnidirectional content processing pipeline is illustrated in Fig. \ref{fig1}. First, multiple viewports are captured by the camera array, and then stitched onto the sphere for covering the entire field of view (FoV). Second, for ease of encoding, transmission and storage, the 3D OIs are projected to ERP format. Here, different coding artifacts would be introduced in both symmetric and asymmetric manners. Finally, after the server storage and transmission, the back projection converting the ERP format to spherical surface is performed and viewport rendering is used to display the watched scene in HMD. Such process involves the unique characteristics of 3D OIs, which should consider the perceptual combination of 3D and omnidirectional content. Additionally, different from ordinary 2D content, a special dimension in 3D vision is depth perception. Therefore, an effective DQA measure for 3D OIs that jointly considers the key factors of 3D depth perception and omnidirectional properties is highly demanded.

Fig. \ref{fig2} depicts examples of three 3D OIs with left and right ERP images. Both left and right views have the same JPEG compression level separately. We find that the image quality values are very similar with each other, while the depth quality values significantly increase with more depth perception. Besides, the overall QoE is comprehensively determined by image quality and depth quality.

Motivated by the above analyses, to fill the gaps, we propose a perceptual depth quality measure that not only handles the depth quality prediction of 3D OIs, but also performs well for single-viewport, i.e., traditional 3D images. Our DQI method adopts the interocular discrepancy statistics of local viewports or regions in various frequency subbands with discriminative depth perception information. Moreover, by considering the perceptual HVS characteristics, color decomposition is used to generate different color channels for interocular discrepancy maps. Experiments on 3D OIQA and 3D IQA subjective databases verify the effectiveness of the proposed DQI and its technical components. In addition, we demonstrate that combining our DQI with image quality features can notably improve the performance results of overall QoE measure.

\section{Proposed Depth Quality Index}
\begin{figure*}[t]
	\centering
	\includegraphics[width=18cm]{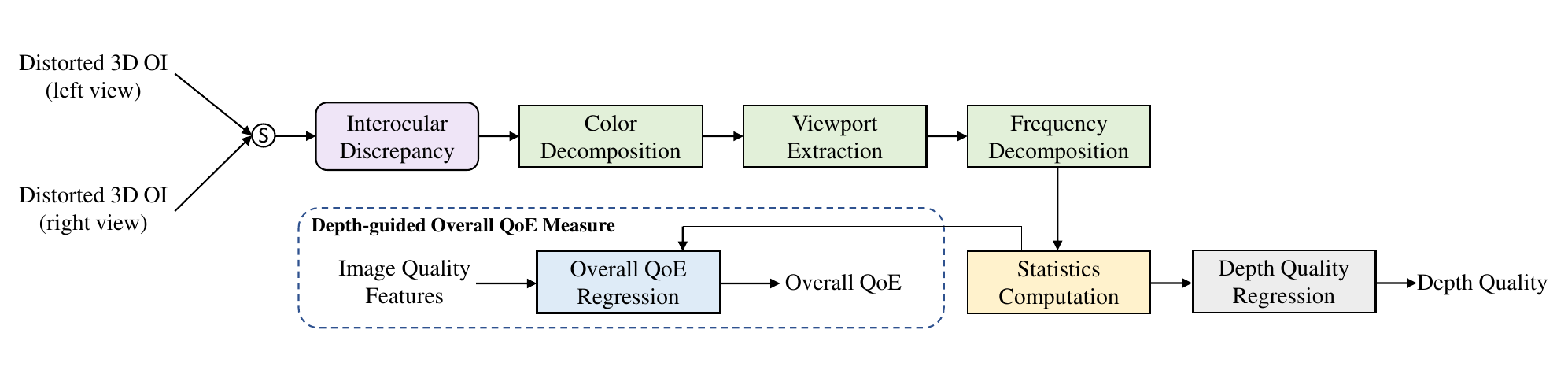}
	\caption{Diagram of the proposed depth quality measure and the depth-guided overall QoE measure, where S represents subtraction operation.}
	\label{fig3}
\end{figure*}

Fig. \ref{fig3} shows the proposed DQI and the depth-guided overall QoE measure. First, the interocular discrepancy of left and right views can effectively reflect depth information. Second, we use color decomposition to generate the visual signals in perceptual color space. Third, based on the observations of resulted luminance and chroma information, we adaptively extract local viewports and then decompose them into the frequency domain. Finally, interocular discrepancy statistics are adopted to be fed into the depth quality regressor which can predict the depth quality. Additionally, our proposed DQI can also be integrated with classical image quality features to produce the overall QoE.

\begin{figure*}[t]
	\centering
	\includegraphics[width=18cm]{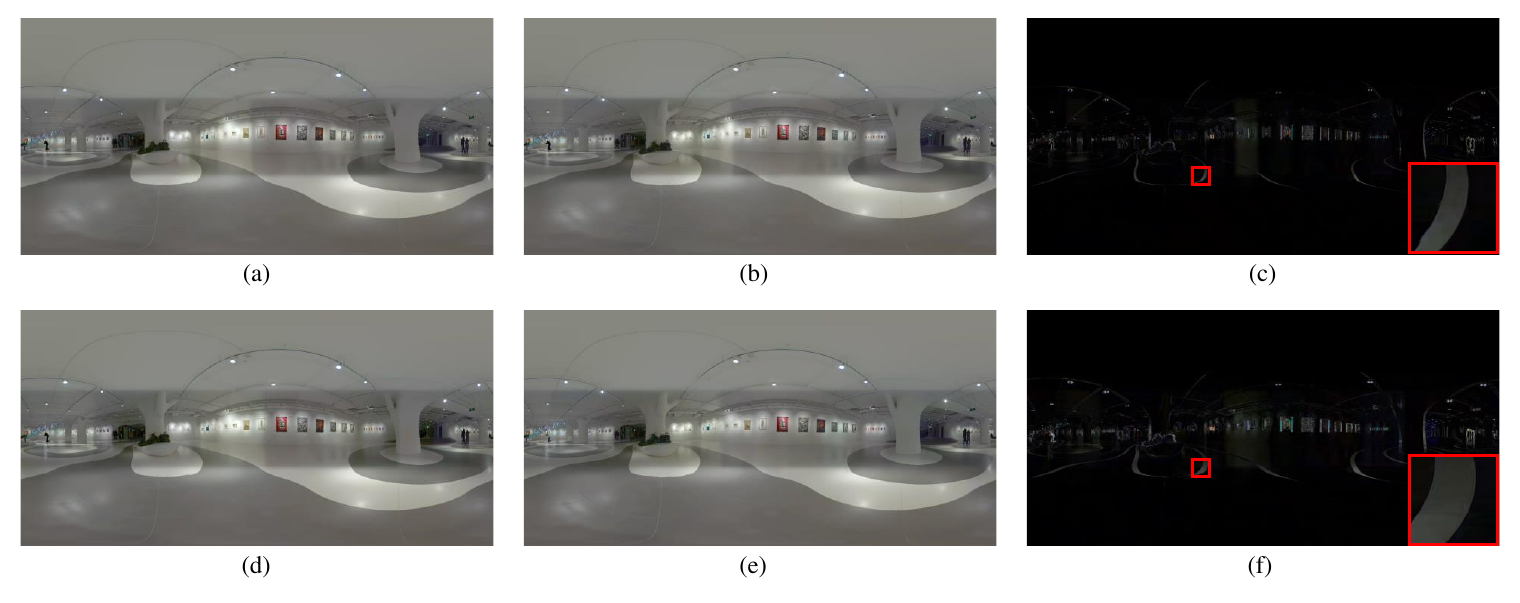}
	\caption{Interocular discrepancy maps with different depth levels. (a) Left view with medium disparity; (b) Right view with medium disparity; (c) Interocular discrepancy map of (a) and (b); (d) Left view with large disparity; (e) Right view with large disparity; (f) Interocular discrepancy map of (d) and (e).}
	\label{fig4}
\end{figure*}

\subsection{Interocular Discrepancy}
With our two eyes viewing the physical world from a slightly different visual perspective, we can feel the sense of depth, which provides the basis of stereopsis \cite{kingdom2012binocular}. If the camera baseline of left and right views becomes larger, the binocular depth level increases, and vice versa. Moreover, based on some research works on psychophysics and neuroscience, the capability of the HVS to match the corresponding features in left and views determines depth perception \cite{ohzawa1990stereoscopic,neri1999probing,parker2007binocular,henriksen2016disparity}. Thus, to reflect different depth information, we here compute the interocular discrepancy as follows:

\begin{equation}
\centering
D=\left|I_{d l}-I_{d r}\right|,
\end{equation}
where $I_{dl}$ and $I_{dr}$ denote left and right view images.

We demonstrate the interocular discrepancy maps with various depth levels in Fig. \ref{fig4}.  As shown in this figure, the first and second rows are two stereopairs with the same compression type and degree. However, (a) and (b), i.e., the 3D OI in the first row has medium disparity, while (d) and (e), i.e., the 3D OI in the second row indicates large disparity. (c) and (f) illustrate the computed interocular discrepancy maps for the corresponding left and right views. We can observe that the interocular discrepancy maps with different depth levels behave differently. Specifically, as seen from the red bounding box, the one that has large disparity delivers more bright regions. Based on this observation, we believe the interocular discrepancy would be a good depth quality indicator.

\begin{figure*}[t]
	\centering
	\includegraphics[width=18cm]{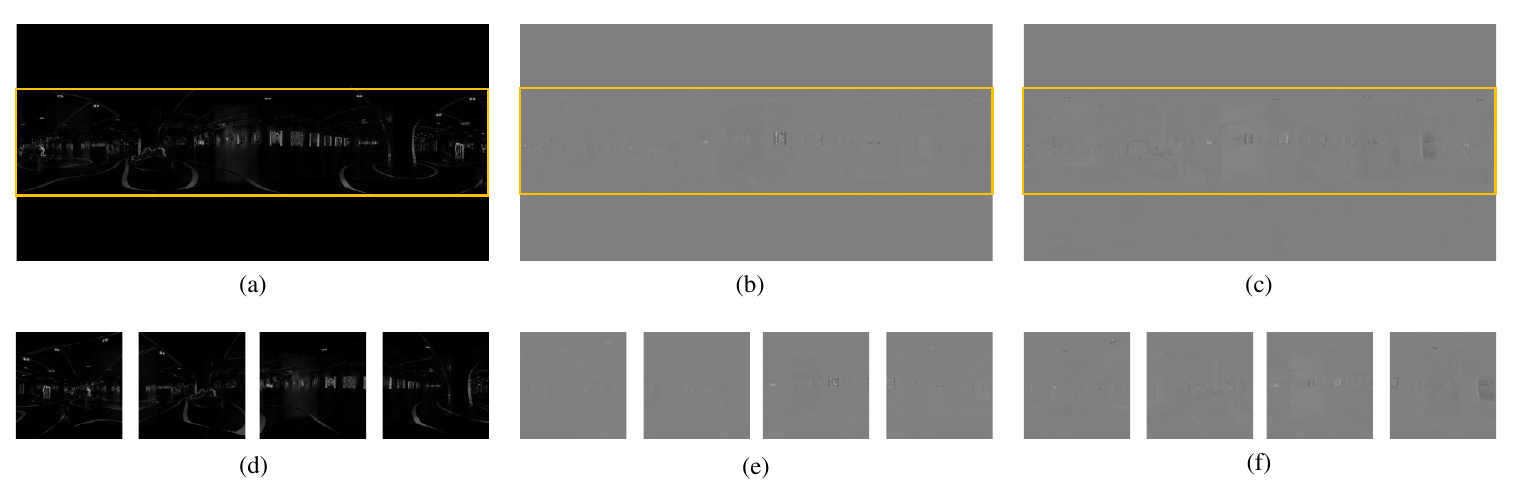}
	\caption{Demonstration of color decomposition and viewport selection for Fig. \ref{fig4} (f). (a-c) Luminance and two chroma components in LAB color space, respectively; (d-f) The corresponding viewports extracted from (a-c).}
	\label{fig5}
\end{figure*}

\begin{figure*}[t]
	\centering
	\includegraphics[width=18cm]{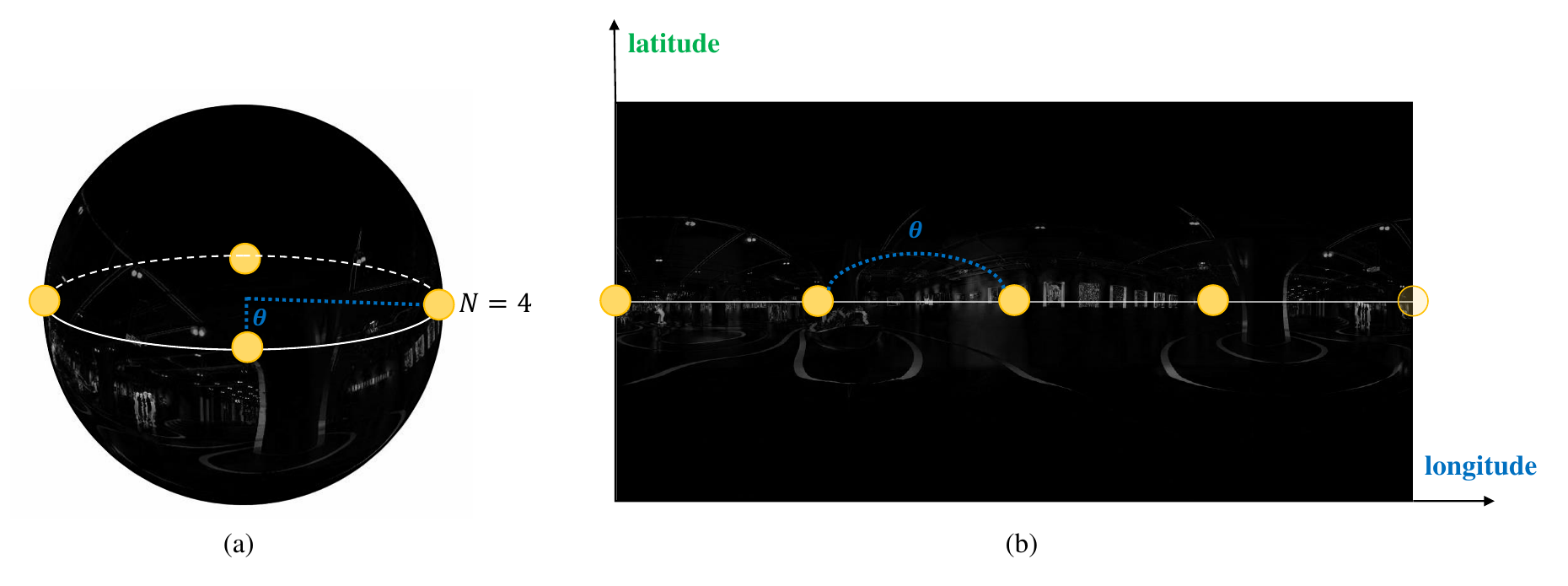}
	\caption{Specific viewport selection process on the sphere and ERP plane. Here we take the luminance component of Fig. \ref{fig5} (a) as an example. (a) Sampling viewports on the sphere; (b) Sampling viewports on the ERP plane.}
	\label{fig6}
\end{figure*}

\subsection{Color Decomposition and Viewport Selection}

\subsubsection{Color Decomposition}
Humans perceive vast spaces of color. As significant perceptual information of the HVS, color visual cues help the human brain to better understand the physical world \cite{boynton1990human}. Many research works have demonstrated the importance of both chroma and luminance on perceptual image quality \cite{preiss2014color,lee2015towards,lee2016toward,ghadiyaram2017perceptual}, but none of them explore this for evaluating the visual quality of 3D OIs. Therefore, we attempt to disentangle the interocular discrepancy map into luminance and chroma components as:

\begin{equation}
\centering
D \sim\left[D_{l}, D_{a}, D_{b}\right],
\end{equation}
where ${D}_l$ is the luminance component, ${D}_a$ and ${D}_b$ are two chroma components. $\sim$ indicates the color decomposition operator.

An example is shown in Fig. \ref{fig5} (a-c). We can see that there exist different visual appearances in these components. Besides, most discriminative information concentrates near the equator. Thus, we choose to extract multiple viewports from the equator of 3D OIs.

\subsubsection{Viewport Selection}
As stated in the section of color decomposition, the selected viewports from the equator of 3D OIs are extracted. In Fig. \ref{fig6}, we give an example of luminance component to show the specific viewport selection process on the sphere and ERP plane, respectively. It should be noted that the adaptive viewport selection process of the other two chroma components is the same as that of the luminance component. In this way, we not only extract multiple viewports containing discriminative depth information, but also avoid the geometric deformation problem.
Suppose that we sample $N$ viewports, the interval between adjacent viewports is calculated by:

\begin{equation}
\theta=\frac{360^{\circ}}{N}.
\end{equation}
Through the viewport selection process, we can obtain multiple viewports denoted by ${D}_{ln}$, ${D}_{an}$ and ${D}_{bn}$ ($n=1, 2, ..., N$) for the luminance and two chroma components, respectively. In our experiments, we adopt $N=4$ for simplification.

\begin{figure*}[t]
	\centering
	\includegraphics[width=18cm]{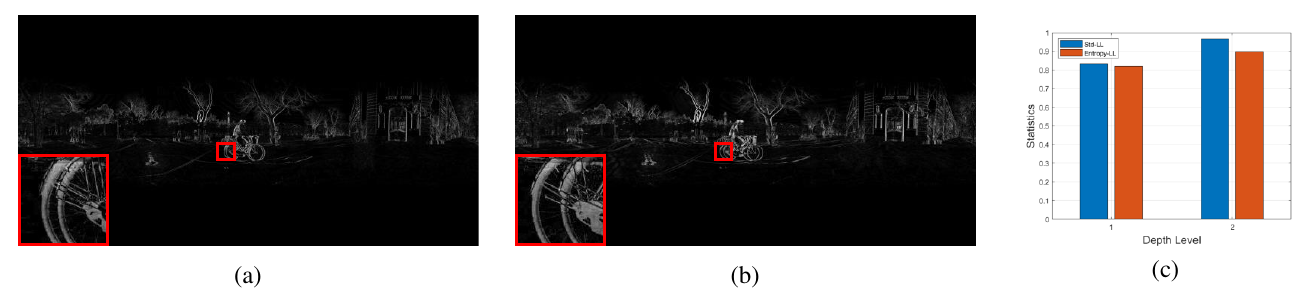}
	\caption{The histograms of statistics for 3D OIs with the same distortion but different depth levels. (a) Luminance component of interocular discrepancy map from 3D OI with medium depth level; (b) Luminance component of interocular discrepancy map from 3D OI with large depth level; (c) The corresponding histograms of statistics for (a) and (b). Note that we omit the zero disparity images since the pixels in interocular discrepancy maps of these images are almost in black.}
	\label{fig7}
\end{figure*}

\subsection{Frequency Decomposition}
Based on the frequency independent mechanism in the human brain, neurons are commonly stimulus selective \cite{heitger1992simulation}. Therefore, image signals are preferred to be treated in the transform domain with different frequencies, instead of directly being used in the original form of entire image. Among lots of transforms, the wavelet transform shows a promising correlation with the HVS \cite{mallat1989multifrequency}. In addition, motivated by the advantages of discrete Haar wavelet transform (DHWT) in perceptual quality modeling \cite{reisenhofer2018haar,zhu2019multi}, here we choose it to decompose the extracted viewports into subbands. Note that since adding more decomposition levels has no significant performance improvement, we use the single level DHWT which can save computational complexity.






Let $W$ be the DHWT matrix and we then can obtain the decomposed viewports in the Haar wavelet domain. Here, we convert luminance viewports as an example:

\begin{equation}
W D_{l n} W^{T}=\left[\begin{array}{c}
\widetilde{D}_{l n}^{L L} \widetilde{D}_{l n}^{H L} \\
\widetilde{D}_{l n}^{L H} \widetilde{D}_{l n}^{H H}
\end{array}\right],
\end{equation}
where the $LL$, $HL$, $LH$ and $HH$ represent subbands with low or high frequency along horizontal or vertical direction. We define the aggregation of luminance subbands after frequency decomposition as $\widetilde{D}_{l n}^{s}, s \in\{L L, H L, L H, H H\}$. Similarly, we can obtain the sets of two chroma subbands in the same way as the luminance, denoted by $\widetilde{D}_{a n}^{s}$ and $\widetilde{D}_{b n}^{s}$.

As shown in Fig. \ref{fig7}, we illustrate the histograms of statistics for 3D OIs with the same distortion but different depth levels. Here, we take the $LL$ subband of luminance component as an example. From this figure, we find that the histograms of statistics for interocular discrepancy can effectively distinguish various depth quality degrees.

\subsection{Statistics for Quality Regression}
Extensive HVS studies have been conducted to demonstrate that the depth perception reveals global properties of visual stimulus \cite{read2005early}. Moreover, many quality assessment works have verified the effectiveness of statistical features \cite{mittal2012no,fang2014no,li2016blind}. Based on the observations of various interocular discrepancy maps, we exploit the representative moment (i.e., standard deviation) and entropy intensity in our framework.

Specifically, the standard deviation of luminance subbands is calculated by:

\begin{equation}
d\left(\widetilde{D}_{l n}^{s}\right)=\sqrt{\mathcal{M}\left[\left(\widetilde{D}_{l n}^{s}-\mathcal{M}\left(\widetilde{D}_{l n}^{s}\right)\right)^{2}\right]},
\end{equation}
where $\mathcal{M}$ denotes the sample mean operator. Besides, the entropy intensity of luminance subbands can be estimated as:

\begin{equation}
e\left(\widetilde{D}_{l n}^{s}\right)=-\sum_{i} p_{i}^{s}\left(\widetilde{D}_{l n}^{s}\right) \log _{2} p_{i}^{s}\left(\widetilde{D}_{l n}^{s}\right),
\end{equation}
where $p_{i}^{s}$ is the probability of the image pixel equaling $i$ in decomposed subbands, which is computed as follows:

\begin{equation}
p_{i}^{s}=\frac{K_{i}^{s}}{K},
\end{equation}
where $K_{i}^{s}$ represents the number of pixels equaling to $i$ in decomposed subbands and $K$ is the total number of pixels.

After obtaining the statistics of luminance subbands, we average them across different viewports by:

\begin{equation}
d\left(\widetilde{D}_{l}^{s}\right)=\frac{1}{N} \sum_{n=1}^{N} d\left(\widetilde{D}_{l n}^{s}\right),
\end{equation}

\begin{equation}
e\left(\widetilde{D}_{l}^{s}\right)=\frac{1}{N} \sum_{n=1}^{N} e\left(\widetilde{D}_{l n}^{s}\right),
\end{equation}

By repeating the same operations as luminance channel $l$ in Eq. (5-9), the statistics for two chroma components can be achieved, which are denoted by $d\left(\widetilde{D}_{a}^{s}\right)$, $e\left(\widetilde{D}_{a}^{s}\right)$ for chroma channel $a$ and $d\left(\widetilde{D}_{b}^{s}\right)$, $e\left(\widetilde{D}_{b}^{s}\right)$ for chroma channel $b$. Then, the final statistical feature for depth quality estimation is composed by:

\begin{equation}
\begin{array}{r}
F_{Depth}=\left[d\left(\widetilde{D}_{l}^{s}\right), d\left(\widetilde{D}_{a}^{s}\right), d\left(\widetilde{D}_{b}^{s}\right)\right., \\
\left.e\left(\widetilde{D}_{l}^{s}\right), e\left(\widetilde{D}_{a}^{s}\right), e\left(\widetilde{D}_{b}^{s}\right)\right],
\end{array}
\end{equation}

Finally, the quality index of our proposed DQI is obtained by the following support vector regression (SVR) \cite{scholkopf2000new} mapping:

\begin{equation}
Q_{Depth}=f\left(F_{Depth}\right),
\end{equation}
where $f(\cdot)$ indicates the SVR function.

\subsection{Extended Overall Quality Prediction}
As an effective DQA method, our proposed depth quality measure should have the capability to assist overall QoE assessment. Thus, by combining with existing image quality features, we extend the proposed DQI to overall quality prediction, namely the depth-guided overall QoE measure, as shown in Fig. \ref{fig3}.
In the proposed depth-guided overall QoE measure framework, we choose the simplest PSNR and a representative structural similarity measure (i.e., MS-SSIM) as the image features. To be specific, the image features of left and right local viewports are calculated as follows:

\begin{equation}
q\left(V_{d l}\right)=\Psi\left(V_{o l}, V_{d l}\right),
\end{equation}

\begin{equation}
q\left(V_{d r}\right)=\Psi\left(V_{o r}, V_{d r}\right),
\end{equation}
where $\Psi$ denotes the local PSNR or MS-SSIM. $V_{o l}$ and $V_{o r}$ are original local viewports, while $V_{d l}$ and $V_{d r}$ are the corresponding distorted local viewports located at the same spatial position. It should be noted that the viewport selection is followed the same way described in Section III-B. Then, the final feature for overall quality prediction is constituted as:

\begin{equation}
F_{\text {Overall }}=\left[q\left(V_{d l}\right), q\left(V_{d r}\right), F_{\text {Depth }}\right].
\end{equation}
Similarly, we can obtain the predicted overall QoE score by the SVR function:

\begin{equation}
Q_{\text {Overall }}=f\left(F_{\text {Overall }}\right).
\end{equation}
With the predicted quality scores and ground-truth subjective ratings, we can compute their correlation as the ultimate performance measurement.

\section{Experimental Results}
In this section, we first introduce the experimental settings including the test databases and criteria. Then, we examine the accuracy and validity of our proposed DQI from three aspects: 1) test the performance of the proposed method for 3D OIQA; 2) simplify the viewport selection process and test the proposed DQI for 3D IQA; 3) extend our proposed depth quality measure to overall QoE prediction and verify its performance. In addition, through the whole experiments, different distortion types and the ablation studies of algorithm components are also considered to further demonstrate our proposed DQI.

\subsection{Evaluation Protocols}
In order to evaluate and compare the performance of our method with state-of-the-arts, we conduct experiments on the SOLID and Waterloo 3D Depth databases. An introduction of them is shown below:

\begin{itemize}
  \item The SOLID database \cite{xu2018subjective} includes 276 distorted 3D OIs, in which 84 images are symmetrically distorted and 192 images are asymmetrically distorted. They are impaired from 6 original reference images, involving two distortion types (i.e., JPEG compression and BPG compression) and three depth levels (i.e., zero, medium and large disparity). All images in this database are with the resolution of $8192 \times 4096$ for single view and stored in the ERP format. Each 3D OI is associated with three labels, namely image quality, depth perception quality and overall quality. These labels are defined as mean opinion scores (MOSs) ranging from 1 to 5, where higher MOSs indicate better quality. It is worth noting that this database is the only 3D OIQA database that provides depth quality.
  \item The Waterloo-IVC 3D Depth database \cite{wang2015depth} originates from 6 pristine texture contents, i.e., Bark, Brick, Flowers, Food, Grass and Water. By considering two depth polarizations and six depth levels, 72 pristine 3D images can be produced. They are degraded by three distortion types containing additive white Gaussian noise, Gaussian blur and JPEG compression. The degradation process can totally result in 1,296 true distorted 3D images, including either inner or outer stereopairs. The image resolution of single view is $480 \times 360$. Each distorted 3D image has a depth perception difficulty index (DPDI) as the ground-truth label. Note that higher DPDI represents lower depth perception.
\end{itemize}

For performance comparisons in our experiments, we adopt three commonly used criteria which are Spearman rank-order correlation coefficient (SROCC), Kendall rank-order correlation coefficient (KROCC), and Pearson linear correlation coefficient (PLCC) \cite{shi2019no,zhou2020tensor}. The definitions of these criteria are depicted as follows:

\begin{equation}
S R O C C=1-\frac{6 \sum_{t=1}^{T} k_{t}^{2}}{T\left(T^{2}-1\right)},
\end{equation}

\begin{equation}
K R O C C=\frac{2\left(P_{c}-P_{d}\right)}{T(T-1)},
\end{equation}

\begin{equation}
P L C C=\frac{\sum_{t=1}^{T}\left(g_{t}-\bar{g}\right)\left(o_{t}-\bar{o}\right)}{\sqrt{\sum_{t=1}^{T}\left(g_{t}-\bar{g}\right)^{2}\left(o_{t}-\bar{o}\right)^{2}}},
\end{equation}
where $T$ is the size of testing set, and $k_t$ represents the rank difference between the subjective and objective scores for the $t$-th image. Moreover, $P_c$ and $P_d$ indicate the numbers of concordant and discordant pairs. In addition, $g_t$ and $o_t$ denote the $t$-th subjective score and mapped objective score after nonlinear regression. The mean of all $g_t$ and $o_t$ are $\bar{g}$ and $\bar{o}$, respectively. Among these criteria, SROCC and KROCC reflect the prediction monotonicity, while PLCC reveals the prediction linearity. An excellent objective quality assessment method is expected to obtain SROCC, KROCC and PLCC close to 1.

\renewcommand\arraystretch{1.0}
\begin{table}[t]
	\centering
	\scriptsize
	\caption{Performance comparison of depth quality prediction on the SOLID database, where DQI- means using traditional non-uniform viewport selection method.}
	\begin{tabular}{|c|c|c|c|c|}
		\hline
		Types & Methods & SROCC & KROCC & PLCC \\ \hline
		\multirow{12}{*}{2D IQA} & PSNR &0.0837 &0.0596 &0.0980 \\
		& SSIM \cite{wang2004image} &0.1296 &0.0884 &0.1119 \\
		& MS-SSIM \cite{wang2003multiscale} &0.0924 &0.0656 &0.0878 \\
		& FSIM \cite{zhang2011fsim} &0.1242 &0.0850 &0.1207 \\
        & BRISQUE \cite{mittal2012no} &0.1768 &0.1247 &0.1681 \\
        & NIQE \cite{mittal2012making} &0.1461 &0.1032 &0.1408 \\ 
        & LPSI \cite{wu2015highly} &0.1644 &0.1156 &0.1524 \\ 
        & dipIQ \cite{ma2017dipiq} &0.0482 &0.0339 &0.0731 \\ 
        & MEON \cite{ma2017end} &0.0988 &0.0671 &0.1362 \\ 
        & CNNIQA \cite{kang2014convolutional} &0.0436 &0.0302 &0.2560 \\ 
        & TRES \cite{golestaneh2022no} &0.0422 &0.0300 &0.0730 \\ 
        & CLIPIQA \cite{wang2023exploring} &0.0375 &0.0218 &0.0477 \\ \hline
		\multirow{4}{*}{2D OIQA} & S-PSNR \cite{yu2015framework} &0.0753 &0.0543 &0.0950 \\
        & WS-PSNR \cite{sun2017weighted} &0.0752 &0.0541 &0.0921 \\
        & CPP-PSNR \cite{zakharchenko2016quality} &0.0754 &0.0547 &0.0946 \\
        & MFILGN \cite{zhou2021no} &0.0138 &0.0101 &0.0533 \\ \hline
		\multirow{3}{*}{3D IQA} & Cyclopean \cite{chen2013full} &0.0513 &0.0327 &0.1559 \\
		& Weighted SSIM \cite{wang2015quality} &0.1248 &0.0854 &0.1101 \\
		& Weighted FSIM \cite{wang2015quality} &0.1218 &0.0808 &0.1214 \\ \hline
		\multirow{3}{*}{DQA} & DPQM \cite{chen2017blind} &0.5588 &0.4001 &0.6524 \\
		& Proposed DQI- &0.9222 &0.7673 &0.9434 \\
		& Proposed DQI &\textbf{0.9299} &\textbf{0.7814} &\textbf{0.9482} \\ \hline
	\end{tabular}
\label{table1}
\end{table}

In the experiments, each database is randomly divided into 80\%-20\% for training and testing. We perform 1,000 iteration times and the median values are regarded as the final measurement. As recommended by the video quality experts group (VQEG) \cite{antkowiak2000final}, a nonlinear mapping should be used before calculating the performance by PLCC. Specifically, we employ a monotonic logistic regression function as:

\begin{equation}
Q^{\prime}(u)=\beta_{1}\left(\frac{1}{2}-\frac{1}{1+e^{\beta_{2}\left(u-\beta_{3}\right)}}\right)+\beta_{4} u+\beta_{5},
\end{equation}
where $u$ and $Q^{\prime}(u)$ are the raw objective quality scores and regressed scores after the nonlinear mapping. $\left\{\beta_{j} \mid j=1,2,3,4,5\right\}$ denote five parameters to be fitted.

\renewcommand\arraystretch{1.0}
\begin{table}[t]
	\centering
	\caption{Performance results of the proposed DQI for different distortion types.}
	\begin{tabular}{|c|c|c|c|}
		\hline
		Distortion Types & SROCC & KROCC & PLCC \\ \hline
		BPG &0.9055 &0.7498 &0.9241 \\ \hline
        JPEG &0.9249 &0.7850 &0.9666 \\ \hline
	\end{tabular}
\label{table2}
\end{table}

\renewcommand\arraystretch{1.0}
\begin{table}[t]
	\centering
	\caption{Performance results of the proposed DQI for symmetrically and asymmetrically distorted 3D OIs.}
	\begin{tabular}{|c|c|c|c|}
		\hline
		Distortions & SROCC & KROCC & PLCC \\ \hline
		Symmetric &0.8472 &0.6982 &0.9098 \\ \hline
        Asymmetric &0.8700 &0.7064 &0.9167 \\ \hline
	\end{tabular}
\label{table3}
\end{table}

\subsection{Performance for Depth Quality Measure of 3D OIQA}
To demonstrate the effectiveness of our proposed method, we conduct experiments on the 3D OIQA database (i.e., SOLID) and compare the DQI with state-of-the-art quality assessment models: 1) classical full-reference 2D IQA approaches including the PSNR, SSIM \cite{wang2004image}, MS-SSIM \cite{wang2003multiscale} and FSIM \cite{zhang2011fsim} as well as no-reference 2D IQA methods consisting of BRISQUE \cite{mittal2012no}, NIQE \cite{mittal2012making}, LPSI \cite{wu2015highly}, dipIQ \cite{ma2017dipiq}, MEON \cite{ma2017end}, CNNIQA \cite{kang2014convolutional}, TRES \cite{golestaneh2022no}, and CLIPIQA \cite{wang2023exploring}; 2) four 2D OIQA metrics, namely S-PSNR\cite{yu2015framework}, WS-PSNR \cite{sun2017weighted}, CPP-PSNR \cite{zakharchenko2016quality}, and MFILGN \cite{zhou2021no}; 3) typical 3D IQA algorithms containing Cyclopean \cite{chen2013full}, Weighted SSIM \cite{wang2015quality} and Weighted FSIM \cite{wang2015quality}; 4) the DQA method called DPQM \cite{chen2017blind}, which is a depth quality measure designed for 3D images. It should be noted that there have been no specifically designed quality metrics for 3D OIs so far.

The performance comparison results of depth quality prediction are reported in TABLE \ref{table1}, where the best performance values are highlighted in bold. From this table, we can find that traditional visual quality assessment models, including 2D IQA, 2D OIQA and 3D IQA methods, fail to evaluate depth quality. This is reasonable because they are designed for assessing image quality rather than depth perception. The image quality targets the perceived quality of pictures, while the depth quality means the ability to deliver an enhanced sensation of depth \cite{zhou20163d}.

Besides, DPQM is a depth quality metric that performs better than those IQA models. However, only 3D images are considered in DPQM, which is not suitable for 3D OIs. To the best of our knowledge, our proposed method is the first one that specifically designs for 3D OIs. The difference between the proposed DQI- and DQI lies in the viewport selection. That is, DQI- adopts the non-uniform viewport sampling method as \cite{xu2019quality,chen2020stereoscopic} which results in six viewports (i.e., four on the equator, one at the north pole and one at the south pole), while our proposed DQI is based on the adaptive viewport selection method described in Section III-B. It can be observed that both DQI- and DQI significantly outperform DPQM, demonstrating the advantages of our designed framework. In addition, the proposed DQI is superior to DQI- because the used viewport selection strategy in this work is more likely to reflect the discriminative information of statistical characteristics for interocular discrepancy maps.

\subsection{Validity of Various Distortion Scenarios}
Since there exist different distortion types and symmetric as well as asymmetric distortions are involved in the SOLID database, we test the performance regards to various distortion scenarios.

The performance results of our method for BPG compression and JPEG compression are shown in TABLE \ref{table2}. As can be seen in the table, the proposed DQI delivers promising performance for the two distortion types. Additionally, from the quantitative numbers of our method for symmetrically and asymmetrically distorted 3D OIs in TABLE \ref{table3}, we can find that DQI can handle both symmetric and asymmetric distortions. These results further verify the effectiveness of our proposed DQI under various distortion scenarios. It should be worth noting that the proposed model shows more superior performance regarding asymmetric distortion. This may be because the 3D OIs with asymmetric distortion have more explicit differences in depth perception.

\renewcommand\arraystretch{1.0}
\begin{table}[t]
	\centering
	\caption{Performance evaluation of the proposed DQI for different interocular discrepancy statistics.}
	\begin{tabular}{|c|c|c|c|}
		\hline
		Methods & SROCC & KROCC & PLCC \\ \hline
		Standard Deviation &0.9196 &0.7646 &0.9413 \\ \hline
        Entropy Intensity &0.8952 &0.7324 &0.9276 \\ \hline
        Proposed DQI &\textbf{0.9299} &\textbf{0.7814} &\textbf{0.9482} \\ \hline
	\end{tabular}
\label{table4}
\end{table}

\renewcommand\arraystretch{1.0}
\begin{table}[t]
	\centering
	\caption{Performance evaluation of the proposed DQI for different color decomposition methods.}
	\begin{tabular}{|c|c|c|c|}
		\hline
		Methods & SROCC & KROCC & PLCC \\ \hline
		$HSV$ &0.9028 &0.7402 &0.9296 \\ \hline
        $LAB$ &\textbf{0.9299} &\textbf{0.7814} &\textbf{0.9482} \\ \hline
	\end{tabular}
\label{table5}
\end{table}

\renewcommand\arraystretch{1.0}
\begin{table}[t]
	\centering
	\caption{Performance evaluation of the proposed DQI for various color channels.}
	\begin{tabular}{|c|c|c|c|}
		\hline
		Channels & SROCC & KROCC & PLCC \\ \hline
		$l$ &\textbf{0.8871} &\textbf{0.7242} &\textbf{0.9232} \\ \hline
        $a$ &0.8776 &0.7040 &0.9176 \\ \hline
        $b$ &0.8450 &0.6665 &0.8912 \\ \hline
	\end{tabular}
\label{table6}
\end{table}

\begin{figure*}[t]
	\centering
	\includegraphics[width=18cm]{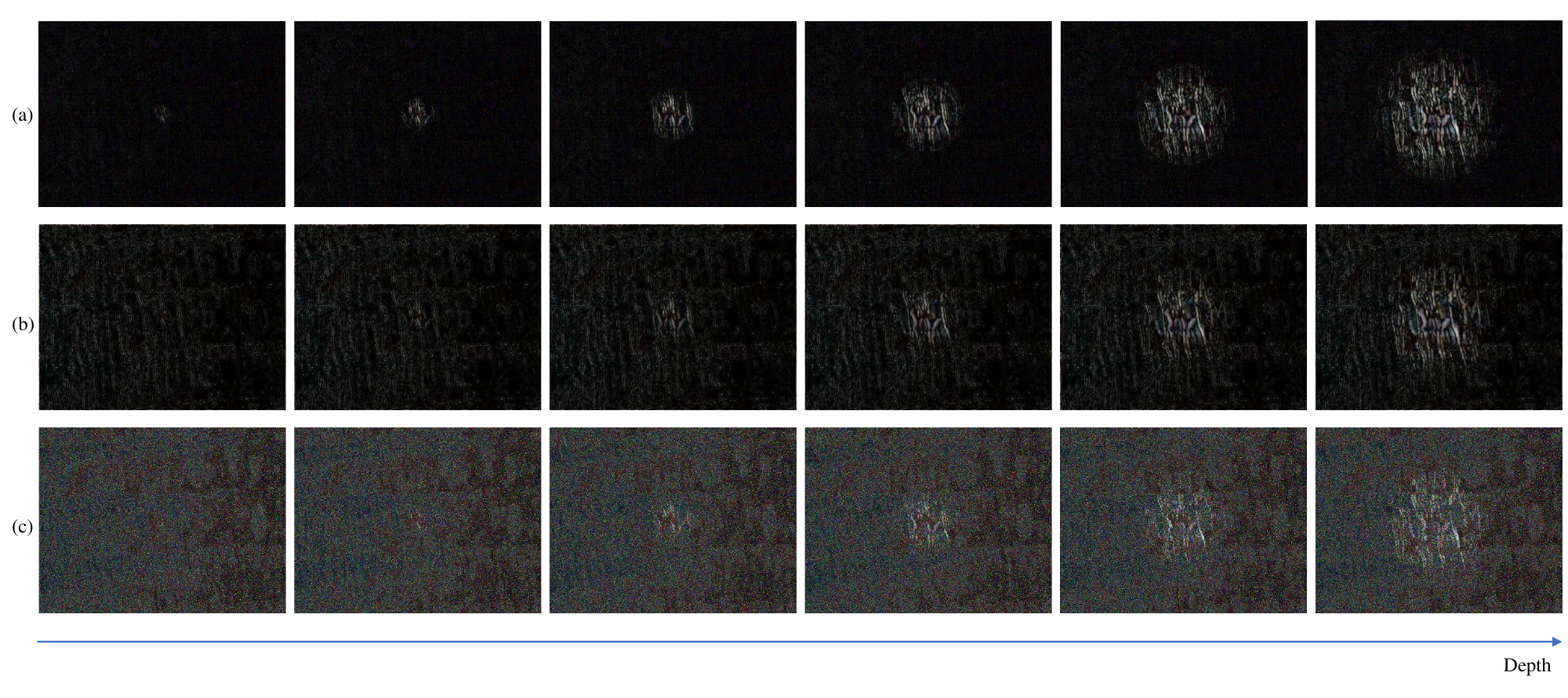}
	\caption{Variation of interocular discrepancy maps with the increase of depth levels. Each row has the same distortion type and degree, but different depth levels. (a) JPEG compression; (b) Gaussian blur; (c) Additive white Gaussian noise.}
	\label{fig8}
\end{figure*}

\subsection{Validity of Different Components}
Considering that our proposed DQI is composed of various components, it is interesting to evaluate the performance for individual component.

First, we test the results of our method about different interocular discrepancy statistics, as illustrated in TABLE \ref{table4}. We can observe that both standard deviation and entropy intensity achieve good performance. Moreover, the combination of them shows the best results. In TABLE \ref{table5} and TABLE \ref{table6}, we then report the performance values of the proposed DQI for various color decomposition methods and color channels. The two tables show that the LAB color space and luminance channel perform better than the HSV color space and the other two chroma channels, respectively. This is mainly due to the prominent role of luminance in the HVS.

\renewcommand\arraystretch{1.0}
\begin{table}[t]
	\centering
	\caption{Performance comparison of depth quality prediction on the Waterloo-IVC 3D depth database, where DQI- means using the entire interocular discrepancy map.}
	\begin{tabular}{|c|c|c|c|}
		\hline
		Methods & SROCC & KROCC & PLCC \\ \hline
		PSNR &0.3600 &0.2712 &0.3239 \\ \hline
        SSIM \cite{wang2004image} &0.3308 &0.2474 &0.3937 \\ \hline
        MS-SSIM \cite{wang2003multiscale} &0.4411 &0.3276 &0.5224 \\ \hline
        FSIM \cite{zhang2011fsim} &0.2925 &0.2195 &0.4342 \\ \hline
        DPDI \cite{wang2016perceptual} &0.8018 &0.6190 &0.7970 \\ \hline
        DPQM \cite{chen2017blind} &0.6355 &0.4631 &0.6445 \\ \hline
		Proposed DQI- &0.7910 &0.6099 &0.8025 \\
		Proposed DQI &\textbf{0.8365} &\textbf{0.6542} &\textbf{0.8526} \\ \hline
	\end{tabular}
\label{table7}
\end{table}

\subsection{Performance for Depth Quality Measure of 3D IQA}
Except for 3D OIs, an intuitive idea is to simplify the viewport selection process to see whether the proposed DQI can still be able to evaluate the depth quality of traditional 3D images or not. One choice is to use the entire interocular discrepancy map, which we denote as DQI-. Here, the question is that does the whole interocular discrepancy map be necessary for depth quality estimation?

We illustrate the variation of interocular discrepancy maps with the increase of depth levels, as depicted in Fig. \ref{fig8}. We take the image content ``Bark'' as an example. Note that each row in this figure has the same distortion type and degree, but different depth levels which increase from left to right. We find that the discriminative information focuses on the map center. Let $X \times Y$ be the resolution of interocular discrepancy maps. For luminance and two chroma channels, we extract the map center as follows:

\begin{equation}
D_{l}{ }^{\prime}=D_{l}\left(\frac{1}{3} Y:\frac{2}{3} Y, \frac{1}{3} X:\frac{2}{3} X\right),
\end{equation}

\begin{equation}
D_{a}{ }^{\prime}=D_{a}\left(\frac{1}{3} Y:\frac{2}{3} Y, \frac{1}{3} X:\frac{2}{3} X\right),
\end{equation}

\begin{equation}
D_{b}{ }^{\prime}=D_{b}\left(\frac{1}{3} Y:\frac{2}{3} Y, \frac{1}{3} X:\frac{2}{3} X\right).
\end{equation}

The above operation is employed to replace the viewport selection process, resulting in the proposed DQI. We conduct experiments on the Waterloo-IVC 3D depth database. TABLE \ref{table7} provides the performance comparison results. It can be seen that our proposed DQI is superior to other state-of-the-art quality assessment models, which validates the potential advantages of our method for evaluating the depth quality of 3D images.

\renewcommand\arraystretch{1.0}
\begin{table}[t]
	\centering
	\scriptsize
	\caption{Performance comparison of overall quality prediction on the SOLID database, where Proposed- means using traditional non-uniform viewport selection method.}
	\begin{tabular}{|c|c|c|c|c|}
		\hline
		Types & Methods & SROCC & KROCC & PLCC \\ \hline
		\multirow{12}{*}{2D IQA} & PSNR &0.5063 &0.3512 &0.5458 \\
		& SSIM \cite{wang2004image} &0.7466 &0.5459 &0.7362 \\
		& MS-SSIM \cite{wang2003multiscale} &0.6247 &0.4447 &0.6376 \\
		& FSIM \cite{zhang2011fsim} &0.7476 &0.5452 &0.7468 \\
        & BRISQUE \cite{mittal2012no} &0.5206 &0.3685 &0.5378 \\
        & NIQE \cite{mittal2012making} &0.5870 &0.4164 &0.6023 \\ 
        & LPSI \cite{wu2015highly} &0.6608 &0.4780 &0.6579 \\ 
        & dipIQ \cite{ma2017dipiq} &0.5387 &0.3803 &0.5410 \\ 
        & MEON \cite{ma2017end} &0.3952 &0.2737 &0.4152 \\ 
        & CNNIQA \cite{kang2014convolutional} &0.1523 &0.0999 &0.2401 \\ 
        & TRES \cite{golestaneh2022no} &0.2246 &0.1542 &0.2587 \\ 
        & CLIPIQA \cite{wang2023exploring} &0.1399 &0.0942 &0.1639 \\ \hline
		\multirow{3}{*}{2D OIQA} & S-PSNR \cite{yu2015framework} &0.4754 &0.3314 &0.5067 \\
        & WS-PSNR \cite{sun2017weighted} &0.4705 &0.3273 &0.4996 \\
        & CPP-PSNR \cite{zakharchenko2016quality} &0.4747 &0.3308 &0.5065 \\
        & MFILGN \cite{zhou2021no} &0.1058 &0.0719 &0.1582 \\ \hline
		\multirow{3}{*}{3D IQA} & Cyclopean \cite{chen2013full} &0.6317 &0.4295 &0.6921 \\
		& Weighted SSIM \cite{wang2015quality} &0.7206 &0.5204 &0.7181 \\
		& Weighted FSIM \cite{wang2015quality} &0.7356 &0.5333 &0.7384 \\ \hline
		\multirow{2}{*}{OQA} & Proposed- &0.9203 &0.7714 &0.9271 \\
		& Proposed &\textbf{0.9301} &\textbf{0.7891} &\textbf{0.9359} \\ \hline
	\end{tabular}
\label{table8}
\end{table}

\subsection{Performance for Overall Quality Measure of 3D OIQA}
Apart from the depth quality measure, we further extend to a depth-guided overall QoE measure. Specifically, we integrate the proposed DQI into existing image quality features. Here, we adopt the well-known MS-SSIM as a representative.

In TABLE \ref{table8}, we show the performance comparison of overall quality prediction on the SOLID database, where OQA means overall quality assessment. The ``Proposed-'' and ``Proposed'' denote the non-uniform viewport sampling method and our proposed viewport selection strategy, respectively. We can see that our extended depth-guided overall QoE measure outperforms state-of-the-art approaches, and the proposed viewport selection method can boost the performance in a sense.

\renewcommand\arraystretch{1.0}
\begin{table}[t]
	\centering
	\caption{Ablation test results of the depth-guided overall QoE measure.}
	\begin{tabular}{|c|c|c|c|}
		\hline
		Methods & SROCC & KROCC & PLCC \\ \hline
		Local PSNR &0.5176 &0.3663 &0.5719 \\ \hline
        Local MS-SSIM \cite{wang2003multiscale} &0.6144 &0.4493 &0.6658 \\ \hline
        Proposed DQI &0.8248 &0.6497 &0.8377 \\ \hline
		Local PSNR+DQI &0.9240 &0.7763 &0.9265 \\
		Local MS-SSIM+DQI &\textbf{0.9301} &\textbf{0.7891} &\textbf{0.9359} \\ \hline
	\end{tabular}
\label{table9}
\end{table}

\subsection{Ablation Study of Overall Quality Prediction}
Since two factors are considered in our depth-guided overall quality measure (i.e., image quality and depth quality), we conduct the ablation study of overall quality prediction by using either image quality features or our proposed DQI.

The performance results are reported in TABLE \ref{table9}, where we use the PSNR and MS-SSIM as image features. It can be observed that even with the simplest PSNR, the proposed DQI can still improve the results. The important boosting effects of our DQI demonstrate its validity for not only depth quality, but also overall QoE.

\section{Conclusion}
In this paper, we propose a DQI method for depth quality assessment based on interocular discrepancy statistics. The proposed model is inspired by the perceptual mechanisms of the HVS, where the interocular discrepancy of the left and right views is used to reflect disparity. Motivated by the color perception in the human brain, we divide the interocular discrepancy into multi-color channels. The key viewports or regions are then selected to perform frequency decomposition and statistics computation, which are on the basis of global depth perception and frequency independent principle in the HVS. Extensive experiments demonstrate that the proposed DQI offers accurate and robust evaluation of depth quality for both 3D omnidirectional images and conventional 3D images. Furthermore, we demonstrate that the proposed DQI can be extended to assess the overall quality by combining with existing image quality features, leading to a depth-guided overall QoE measure. Experimental results show that the proposed method outperforms state-of-the-art quality assessment models. Future work includes the incorporation of deeper visual models such as depth polarization and the development of comprehensive QoE assessment models jointly considering image quality, depth quality and visual discomfort aspects.


%





\ifCLASSOPTIONcaptionsoff
  \newpage
\fi



%
\bibliographystyle{IEEEtran}
\bibliography{tipRefer}



%

%
%
%




\end{document}